%% file: main.tex
\documentclass[sigconf,nonacm]{acmart}

\AtBeginDocument{  }

\setcopyright{none}
\renewcommand\footnotetextcopyrightpermission[1]{}
\usepackage[nameinlink,capitalise]{cleveref}
\usepackage{multirow}
\usepackage{microtype}

\begin{document}

\title{Semantic-Critical Regions for Video-to-Audio Generation}

\author{Yun Fang}
\email{yun.fang@sydney.edu.au}
\affiliation{  \institution{The University of Sydney}
  \city{Sydney}
  \country{Australia}
}

\author{Wei Guo}
\email{wei.guo@sydney.edu.au}
\affiliation{  \institution{The University of Sydney}
  \city{Sydney}
  \country{Australia}
}

\author{Zhuonan Liang}
\email{zhuonan.liang@sydney.edu.au}
\affiliation{  \institution{The University of Sydney}
  \city{Sydney}
  \country{Australia}
}

\author{Heng Wang}
\email{hengwang.vision@gmail.com}
\affiliation{  \institution{The University of Sydney}
  \city{Sydney}
  \country{Australia}
}

\author{Weidong Cai}
\email{tom.cai@sydney.edu.au}
\affiliation{  \institution{The University of Sydney}
  \city{Sydney}
  \country{Australia}
}

\begin{abstract}

In real-world videos, sound events are often associated with localized and fine-grained visual cues, such as object motion, object-part dynamics, contact events, and interactions between objects. Existing video-to-audio (V2A) methods employ implicit conditioning based on global visual representations or explicit control derived from temporal and spatial cues. However, subtle sound-producing and sound-indicative interactions remain insufficiently captured, leaving room for more precise visual conditioning of sound-event semantics and temporal dynamics. 
Motivated by this observation, we refer to regions containing such sound-relevant visual evidence as \textbf{Semantic-Critical Regions (SCRs)} and introduce \textbf{SCR-V2A}, a V2A generation framework that explicitly localizes and encodes SCRs as fine-grained visual guidance for audio generation.
Specifically, we first identify SCRs through sound-related visual localization. We then transform the localized regions into a control representation through adapter-based dual visual encoding, which jointly captures global visual context and fine-grained sound-relevant evidence. The representation is used to condition a downstream audio generation model. 
Experiments on both in-domain and out-of-domain benchmarks demonstrate that SCR-V2A achieves strong performance in distribution matching and audio quality while remaining competitive in cross-modal alignment and requiring few trainable parameters.
These results highlight the value of prioritizing fine-grained, sound-relevant visual regions for V2A generation.

\end{abstract}

\begin{CCSXML}
<ccs2012>
   <concept>
       <concept_id>10002951.10003227.10003251.10003256</concept_id>
       <concept_desc>Information systems~Multimedia content creation</concept_desc>
       <concept_significance>500</concept_significance>
    </concept>
   <concept>
       <concept_id>10010405.10010469.10010475</concept_id>
       <concept_desc>Applied computing~Sound and music computing</concept_desc>
       <concept_significance>500</concept_significance>
    </concept>
 </ccs2012>
\end{CCSXML}

\ccsdesc[500]{Information systems~Multimedia content creation}
\ccsdesc[500]{Applied computing~Sound and music computing}

\maketitle

\input{sections/1_intro}

\input{sections/2_relatedwork}

\input{sections/3_method}

\input{sections/4_expr}

\input{sections/5_conc}

\clearpage \bibliographystyle{ACM-Reference-Format}
\bibliography{main}

\appendix

\clearpage

\twocolumn[
    \centering
    {\fontsize{16pt}{20pt}\selectfont
    \bfseries Semantic-Critical Regions for Video-to-Audio Generation \par}
    \vspace{1em}
    {\fontsize{14pt}{15pt}\selectfont
    \bfseries  Supplementary Materials \par}
    \vspace{1.5em}
]

\input{sections/X_impl_details}

\input{sections/X_expr_ratio}

\input{sections/X_expr_negative_heatmap}

\input{sections/X_human_eval}

\input{sections/X_limitation}

\end{document}

%% file: sections/1_intro.tex
\section{Introduction}\label{sec:1_intro}
Video-to-audio (V2A) generation~\cite{luo2023diff,comunita2024syncfusion,xing2024seeing,xie2024sonicvisionlm,wang2024v2a, VTA-LDM,VideoFoley,Draw-an-Audio,jeong2025read,TA-V2A,MultiFoley,FoleyCrafter} aims to synthesize plausible audio that is semantically and temporally aligned with visual input. Achieving such alignment, however, is challenging because sound events are often associated with fine-grained visual cues that may be subtle, transient, or embedded within complex scene dynamics. In real-world scenes, sounds are produced by, or temporally correlated with, object motion, object-part dynamics, and interactions between objects. For example, speech dynamics can be inferred from lip and facial movements; a bouncing basketball produces sound upon contact with the ground; and guitar sounds vary with finger movements.

\input{figureTex/Fig1_teaserFigure}

Existing V2A methods have explored a range of conditioning approaches to improve semantic and temporal alignment. These methods can broadly be categorized into implicit conditioning based on global visual representations~\cite{iashin2021taming,sheffer2023hear, luo2023diff,CLIPSonic, wang2024v2a} and explicit control derived from temporal~\cite{FoleyCrafter,jeong2025read,VideoFoley,PhysicsDrivenDiff, TiVA, Draw-an-Audio} or spatial visual cues~\cite{Draw-an-Audio,SS2A,RD-FGFS}. Nevertheless, subtle sound-producing and sound-indicative interactions remain insufficiently captured, leaving room for more precise visual conditioning of sound-event semantics and temporal dynamics. 
Together, these observations highlight fine-grained regions associated with sound production as a potentially informative source of visual conditioning for V2A generation.

Motivated by this, we explicitly localize and encode sound-relevant visual evidence to provide more informative control for V2A generation.
As illustrated in \Cref{fig:Fig1_teaserFigure}, we refer to the regions containing such evidence as \textbf{Semantic-Critical Regions (SCRs)}, which provide localized visual cues associated with sound-event semantics and sound production.
To this end, we propose a novel framework, \textbf{SCR-V2A}, which jointly encodes localized SCRs and the original video, and uses the fused representation to condition downstream audio generation. 
Specifically, we first identify SCRs through weakly supervised sound-related visual localization, which predicts sound-event tags and highlights the visual regions that provide discriminative evidence for the predicted events. We then process the original video and the resulting SCR-guided heatmap video through an adapter-based dual visual encoding module, obtaining complementary contextual and localized representations that are fused into a V2A control representation. Finally, the fused representation is incorporated into a control adapter to provide multi-scale visual guidance to a pretrained audio latent diffusion model, while the predicted sound-event tags serve as an auxiliary semantic condition.
Experiments show that SCR-V2A achieves leading performance in distribution matching and audio quality on VGGSound 
, while remaining competitive in cross-modal alignment 
against recent state-of-the-art V2A methods. Its strong performance on the out-of-domain CountixAV test set further demonstrates its robustness and generalization under domain shift.
Notably, SCR-V2A achieves these results while requiring substantially fewer trainable parameters than several large-scale competing methods.

Our contribution can be summarized as follows:

\begin{itemize}
\item We propose SCR-V2A, a novel V2A model that identifies
localized semantic evidence associated with sound produc-
tion and incorporates these spatial cues into an LDM via
an adapter-based conditioning architecture.

\item SCR-V2A achieves leading generation fidelity and remains
competitive in cross-modal alignment on VGGSound with
few trainable parameters, while exhibiting robustness and
generalization under distribution shift on CountixAV.

\item We ablate visual priors for modeling SCRs and show that 
class activation maps (CAMs) effectively capture fine-grained 
visual cues associated with sound-producing events.

\item We further investigate temporal sound-event composition
and sound-source awareness, demonstrating the framework’s
extensibility for temporally structured audio synthesis and
localized reasoning over sound-producing objects.

\end{itemize}

\color{black}

%% file: figureTex/Fig1_teaserFigure.tex
\begin{figure}[t]
  \centering
  \includegraphics[width=0.85\linewidth]{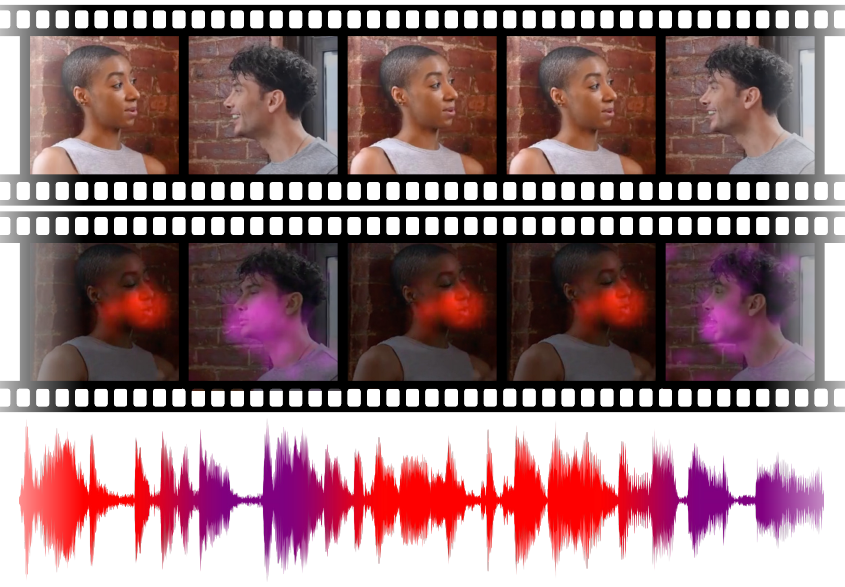}
  \Description{A video sequence of two people speaking is shown as a filmstrip. The corresponding SCR heatmaps highlight the active speaker's mouth in red or purple in each frame, and the aligned audio waveform below uses the same colors to indicate the alternating speakers.}
  \caption{
SCR regions highlight fine-grained visual evidence associated with sound-producing events.
}
  \label{fig:Fig1_teaserFigure}
\end{figure}

%% file: sections/2_relatedwork.tex
\section{Related Work}\label{sec:2_relatedwork}

\subsection{V2A Generation}
\noindent\textbf{Implicit Control.}
Early approaches to V2A generation predominantly rely on implicit conditioning, where visual inputs are encoded into latent representations without explicit structural guidance. 
\cite{iashin2021taming, sheffer2023hear, luo2023diff, CLIPSonic, wang2024v2a} learn direct mappings from visual features to audio signals using generative models, such as GANs and diffusion models. While these methods establish effective frameworks for V2A generation, their reliance on global representations may obscure fine-grained cross-modal correspondences, thereby limiting audio-visual alignment and generation controllability.

\noindent\textbf{Explicit Temporal Control.} 
Studies guide audio generation using explicit temporal or acoustic conditions, including sound-event timestamps~\cite{FoleyCrafter}, audio energy~\cite{jeong2025read}, root-mean-square energy~\cite{VideoFoley}, physics-related parameters~\cite{PhysicsDrivenDiff}, and structured musical plans~\cite{MusicWeaver}. Beyond these conditions, SpecMaskFoley~\cite{SpecMaskFoley} aligns visual features with audio tokens for temporal guidance, while TiVA~\cite{TiVA} predicts audio layouts to model temporal structure. 
Our method focuses on explicit spatial modelling without additional temporal control to isolate the contribution of spatial guidance, while remaining complementary to existing temporal conditioning mechanisms.

\noindent\textbf{Explicit Spatial Control.} 
Parallel to temporal guidance, explicit spatial cues have been explored to improve sound source awareness. Methods such as Draw-an-Audio~\cite{Draw-an-Audio} incorporate visual object masks along with audio attributes (e.g., loudness), while SS2A~\cite{SS2A} focuses on sound source localization. Task-specific approaches like RD-FGFS~\cite{RD-FGFS} further leverage human pose estimation to model motion-induced sounds. 
While these methods provide explicit spatial grounding, they primarily operate at the object level and treats each object entity as a whole. However, the fine-grained visual semantic cues relevant to sound generation may not effectively be captured by such object-level representations.

\input{figureTex/Fig2_method}
\subsection{Visual Localization Cues}
Visual cues provide spatial information for vision and multimodal tasks. Existing approaches use different cues to capture object locations, scene structures, motion patterns, and semantic evidence. 
Representative examples include object bounding boxes for object-level localization~\cite{RCNN,FasterRCNN}, segmentation masks for pixel-wise region representation~\cite{MaskRCNN}, optical flow for motion estimation~\cite{FlowNet,RAFT}, pose-based representations for capturing articulated body structures~\cite{TokenPose,KeyRegionPose}, and transformer-based approaches for modeling spatial feature interactions and highlighting informative visual regions~\cite{ViT,AttnRollout, guo2026collapse}.
These cues encode complementary spatial information and support downstream tasks.
Among them, Class Activation Maps (CAMs) provide weakly supervised localization by highlighting discriminative evidence for category prediction~\cite{zhou2016learning}. Numerous variants~\cite{GradCAM, GradCAM++, ScoreCAM, LayerCAM, EigenCAM} further improve localization quality and interpretability. Unlike explicit cues that often require additional annotations, CAMs can be extracted directly from pretrained recognition models without extra supervision, providing efficient and scalable semantic-aware spatial information.
Prior studies mainly use CAMs for model interpretation, visual explanation, and weakly supervised localization. We instead employ CAMs to model SCRs for V2A tasks, as they identify fine-grained regions contributing to sound-event tag prediction, while continuous activation values encode their relative importance.

\subsection{Adapter-based Conditional Generation}

Adapters~\cite{houlsby2019parameter, stickland2019bert, pfeiffer2020adapterhub, li2021benchmarking, chen2023vision, lin2025ctrl, duan2025unic} enable parameter-efficient adaptation of pretrained models by introducing lightweight trainable modules while preserving the knowledge encoded in the backbone. In conditional diffusion models, they can incorporate additional control signals without fine-tuning the entire generative model~\cite{controlnet, ye2023ip, li2023gligen, mou2024t2i}. Following this paradigm, we use adapters for dual visual encoding and multi-level injection of SCR-based guidance into a pretrained audio diffusion backbone.

%% file: figureTex/Fig2_method.tex
\begin{figure*}[!htbp]
  \centering
  \includegraphics[width=0.95\linewidth]{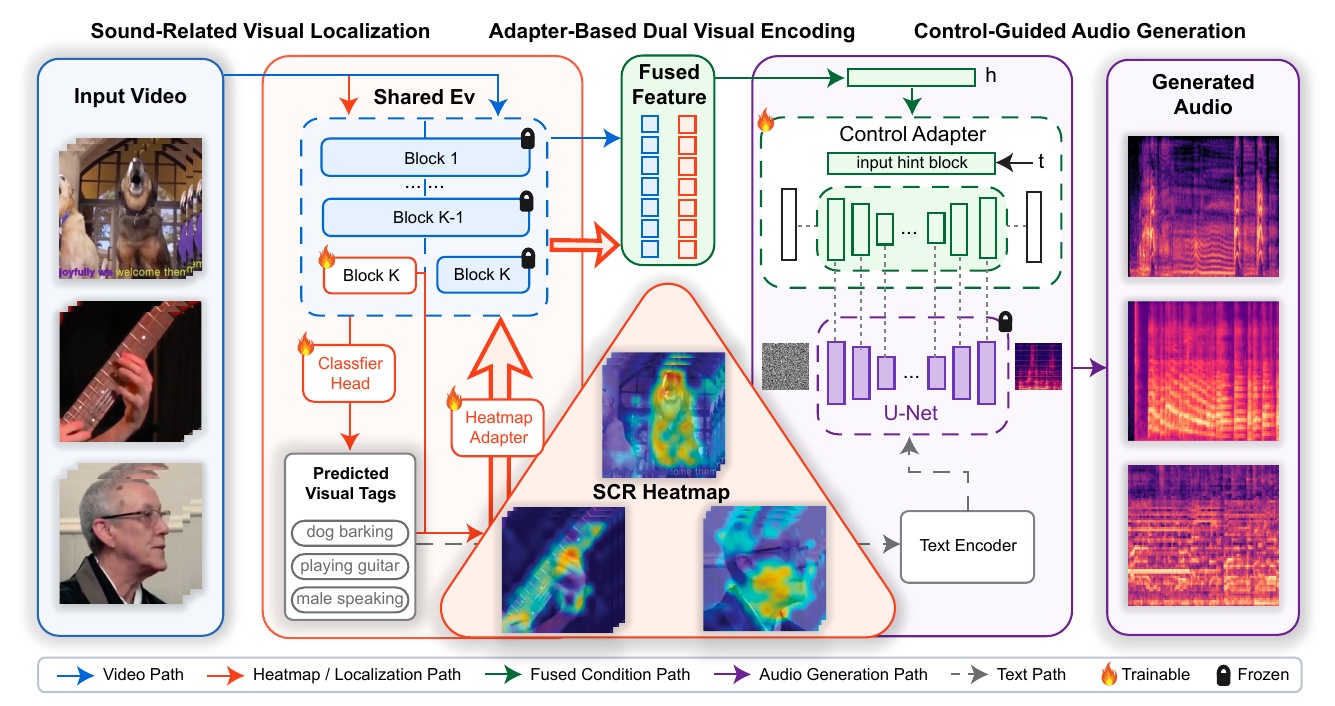}
  \Description{Pipeline of SCR-V2A from input video to generated audio. A shared video encoder predicts sound-related tags and produces SCR heatmaps through a trainable classifier and heatmap adapter. Original-video and heatmap features are fused, passed through a control adapter into a frozen audio diffusion U-Net, and combined with text features to generate an audio spectrogram. Colored arrows distinguish video, localization, fused-condition, audio-generation, and text paths; flame and lock icons mark trainable and frozen components.}
  \caption{Overview of SCR-V2A. Our method formulates Semantic-Critical Regions (SCRs) as the control for V2A generation. The SCRs are identified through Sound-Related Visual Localization, transformed into a control representation via Adapter-Based Dual Visual Encoding, and injected into the diffusion model for Control-Guided Audio Generation.
    }
  \label{fig:Fig2_method}
\end{figure*} 

%% file: sections/3_method.tex
\section{Method}
\label{sec:3_method}

As illustrated in \Cref{fig:Fig2_method}, our framework consists of three stages.
Given videos $V \in \mathbb{R}^{B \times T \times C \times H \times W}$, where $B$, $T$, $C$, and $H \times W$ denote the batch size, number of frames, input channels, and spatial resolution, respectively, we first perform Sound-Related Visual Localization to identify SCRs.
This module predicts sound-event tags $\hat{T} \in [0,1]^{B \times N}$ and produces SCR-guided heatmaps $M \in \mathbb{R}^{B \times T \times C \times H \times W}$.
Second, the SCR heatmaps and original video are encoded by an Adapter-Based Dual Visual Encoding module into visual representations $F_v$ and $F_m$, respectively, and fused through feature concatenation into a V2A control representation $F_f$.
Finally, the Control-Guided Audio Generation module conditions on $F_f$ and the auxiliary tags $\hat{T}$ to generate audio waveforms $\hat{A}$.

\input{figureTex/Fig3_qualitativeResults}

\subsection{Extracting SCRs for V2A Control}
\label{sec:scr_extraction}

Given an input video $V$, we employ a MotionFormer-based encoder $E_v$ to predict sound-event probabilities:

\begin{equation}
\hat{T}
=
\sigma
\left(
C_{\mathrm{cls}}
\left(
E_v(V)
\right)
\right),
\qquad
\hat{T}
\in
[0,1]^{B \times N},
\label{eq:tag_prediction}
\end{equation}

\noindent where $C_{\mathrm{cls}}$ is a lightweight multi-label classification head, $N$ is the number of tags, and $\hat{T}_{b,n}$ denotes the predicted probability of the $n$-th tag for the $b$-th video.
We initialize $E_v$ from the MotionFormer-based model used in SynchFormer~\cite{synchformer,motionformer}, freeze all but its final transformer block, and append $C_{\mathrm{cls}}$ for sound-event recognition. The trainable components are optimized using binary cross-entropy:

\begin{equation}
\mathcal{L}_{\mathrm{cls}}
=
-\frac{1}{BN}
\sum_{b=1}^{B}
\sum_{n=1}^{N}
\left[
T_{b,n}\log \hat{T}_{b,n}
+
\left(1-T_{b,n}\right)
\log\left(1-\hat{T}_{b,n}\right)
\right],
\label{eq:classification_loss}
\end{equation}

\noindent where $\mathbf{T} \in \{0,1\}^{B \times N}$ denotes the ground-truth multi-hot labels. This partial fine-tuning enables the encoder to capture sound-related semantic evidence while largely preserving the visual representations learned during pretraining.
After training, we extract tokens from the first normalization layer of the final transformer block. For each video, we retain at most five predictions above an empirically selected threshold of 0.65, denoted by $\mathcal{S}_b$. After removing special tokens, the remaining patch tokens are reshaped into $X \in \mathbb{R}^{B \times D \times T' \times H' \times W'}$.
For each selected tag $n \in \mathcal{S}_b$, the contribution weight of the $d$-th feature map is obtained by globally averaging the gradients of its score $s_{b,n}$:

\begin{equation}
\alpha_{b,d}^{(n)}
=
\frac{1}{T'H'W'}
\sum_{\tau=1}^{T'}
\sum_{i=1}^{H'}
\sum_{j=1}^{W'}
\frac{\partial s_{b,n}}
{\partial X_{b,d,\tau,i,j}},
\label{eq:scr_weight}
\end{equation}

\noindent where $d \in \{1,\ldots,D\}$ indexes the feature maps, $\tau \in \{1,\ldots,T'\}$ indexes the temporal dimension, and $i \in \{1,\ldots,H'\}$ and $j \in \{1,\ldots,W'\}$ index the two spatial dimensions. The resulting channel-wise weights are then used to obtain a low-resolution class activation map for each selected tag:

\begin{equation}
\tilde{M}_{\mathrm{low},b}^{(n)}
=
\operatorname{ReLU}
\left(
\sum_{d=1}^{D}
\alpha_{b,d}^{(n)} X_{b,d}
\right),
\qquad
\tilde{M}_{\mathrm{low},b}^{(n)}
\in
\mathbb{R}^{T' \times H' \times W'},
\label{eq:scr_response}
\end{equation}

\noindent where $X_{b,d} \in \mathbb{R}^{T' \times H' \times W'}$ denotes the $d$-th feature map of the $b$-th video.
Each response map is then interpolated to the temporal and spatial resolution of the input video and normalized to $[0,1]$:

\begin{equation}
\tilde{M}_{b}^{(n)}
=
\operatorname{Norm}
\left(
\operatorname{Interp}
\left(
\tilde{M}_{\mathrm{low},b}^{(n)};
T,H,W
\right)
\right),
\quad
\tilde{M}_{b}^{(n)}
\in
[0,1]^{T \times 1 \times H \times W}.
\label{eq:grayscale_scr_individual}
\end{equation}

\noindent We then aggregate the response maps of all selected tags by element-wise maximum to obtain a unified SCR map $\tilde{M}$ for each video.
Finally, we colourize the aggregated response map and overlay it on the corresponding RGB frames to construct the SCR-guided heatmap video for subsequent visual encoding:

\begin{equation}
M
=
\lambda V
+
(1-\lambda)
\operatorname{Colorize}
\left(
\tilde{M}
\right),
\qquad
M
\in
\mathbb{R}^{B \times T \times C \times H \times W},
\label{eq:rgb_heatmap}
\end{equation}

\noindent where $\lambda=0.5$ gives equal weight to the RGB and SCR.

\subsection{Learning V2A Control via Visual Encoding}
\label{sec:visual_encoding}

To jointly preserve global visual context and emphasize sound-related semantic evidence, we introduce an Adapter-Based Dual Visual Encoding module to encode the original video and the extracted SCR heatmaps.
The original video is encoded by the video encoder $E_v$ to obtain the global visual representation $F_v$.
Since SCR heatmaps differ from natural RGB videos, they are processed by a dedicated heatmap adapter $A_m$ to reduce the domain mismatch before being passed to the same video encoder. Sharing $E_v$ keeps $F_v$ and $F_m$ in a consistent feature space for subsequent fusion. 
The SCR representation is thus obtained as:

\begin{equation}
F_m
=
E_v
\left(
A_m(M)
\right).
\label{eq:heatmap_feature}
\end{equation}

\noindent This adapter comprises 3D convolutional layers, group normalization, and GELU activations, together with a scaled residual connection controlled by a learnable scalar.
We then concatenate $F_v$ and $F_m$ along the feature dimension to obtain $F_f$, which retains global visual context for realistic audio synthesis while emphasizing sound-relevant visual evidence.

\subsection{Control-Guided Audio Generation}
\label{sec:audio_generation}

Given the fused visual feature $F_f$ and the predicted sound-event tags $\hat{T}$, we generate the corresponding audio using a control adapter $A_{\mathrm{C}}$ integrated with a pretrained audio generator~\cite{audioldm2,liu2023audioldm}. The adapter first transforms $F_f$ into control features compatible with the denoising U-Net, while incorporating the diffusion timestep embedding to account for different denoising stages. Following the standard design in~\cite{controlnet}, the resulting multi-scale control features are injected into the middle block and the corresponding encoder-decoder skip connections, enabling the fused visual condition to guide audio generation across different feature levels. Meanwhile, $\hat{T}$ serves as an auxiliary semantic condition for the LDM. The overall controlled noise prediction is expressed as:
\begin{equation}
    \hat{\boldsymbol{\epsilon}}
    =
    \epsilon_{\theta}
    \left(
        \mathbf{z}_t,
        t,
        \hat{T};
        A_{\mathrm{C}}(F_f,t)
    \right),
\end{equation}
where $\mathbf{z}_t$ is the noisy audio latent at timestep $t$, and
$\epsilon_{\theta}$ denotes the denoising U-Net. After iterative
denoising, the resulting latent is decoded into a mel-spectrogram and
converted into the generated waveform $\hat{\mathbf{A}}$ using the
pretrained decoder and vocoder.
The model is trained using the standard latent diffusion denoising objective:
\begin{equation}
    \mathcal{L}_{\mathrm{diff}}
    =
    \mathbb{E}_{\mathbf{z}_0,t,\boldsymbol{\epsilon}}
    \left[
        \left\|
            \boldsymbol{\epsilon}
            -
            \epsilon_{\theta}
            \left(
                \mathbf{z}_t,
                t,
                \hat{T};
                A_{\mathrm{C}}(F_f,t)
            \right)
        \right\|_2^2
    \right],
\end{equation}
where $\boldsymbol{\epsilon}$ denotes the Gaussian noise added to the audio latent $\mathbf{z}_0$.

\color{black}

%% file: figureTex/Fig3_qualitativeResults.tex
\begin{figure*}[!htbp]
  \centering
  \includegraphics[width=0.98\linewidth]{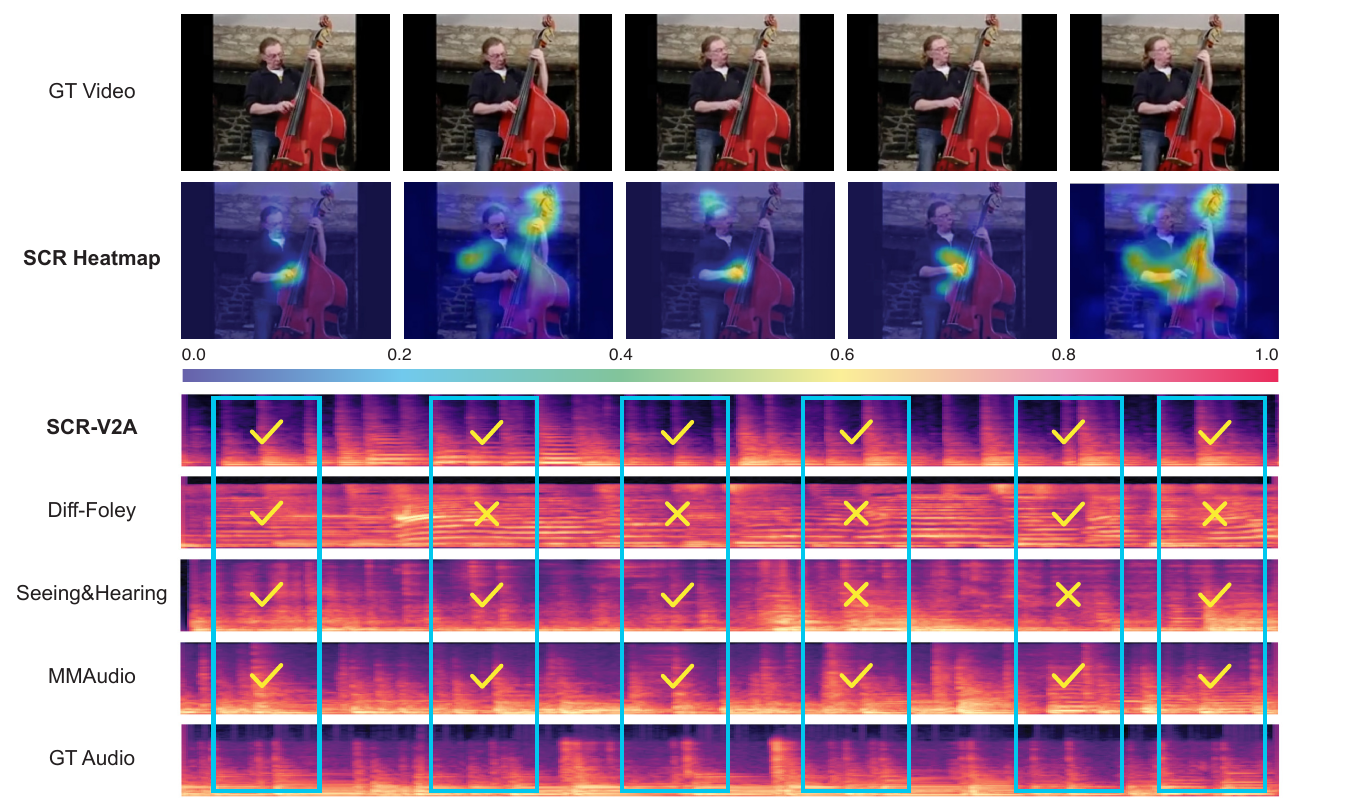}
  \Description{Qualitative comparison for a video of a musician playing a double bass. Five video frames and their SCR heatmaps emphasize the moving hand, bow, face, and instrument. Below, spectrograms from SCR-V2A, Diff-Foley, Seeing and Hearing, MMAudio, and ground-truth audio are aligned in rows. Cyan boxes mark five temporal segments; yellow check marks indicate a detected musical event and crosses indicate a missing event. SCR-V2A contains the event in all five segments and most closely follows the ground-truth temporal pattern.}
  \caption{Qualitative Results. 
  \textcolor{blue}{Blue boxes} highlight representative temporal segments where spectral energy patterns are compared across methods.
  \textcolor{orange}{Yellow marks} indicate whether the sound event is present in each generated audio.
  }
  \label{fig:Fig3_qualitativeResults}
\end{figure*}

%% file: sections/4_expr.tex
\input{tableTex/Tab1_objectiveEval}

\section{Experiment and Results}\label{sec:4_expr}

\subsection{Settings}
\noindent\textbf{Datasets.}
We train SCR-V2A on the VGGSound dataset \cite{VGGSound}, which contains \textasciitilde200K 10-second YouTube videos, covering 310 sound event classes. We follow the official train/test split and reserve 2K videos from the training set for validation.
For cross-dataset evaluation, we further test on Countix-AV \cite{countixav}, a human-action-centric benchmark with dense audio-visual correspondence cues, to assess cross-domain generalisation beyond the VGGSound distribution.

\noindent\textbf{Objective Metrics.} We report objective metrics covering three aspects: distribution matching, audio quality, and cross-modal alignment. Distribution matching is assessed using Fréchet Distance (FD) computed with PANNs~\cite{kong2020panns}, Fréchet Audio Distance (FAD) computed with VGGish~\cite{VGGish}, and Kullback–Leibler (KL) divergence computed from PANNs classifier outputs. Audio quality and diversity are evaluated using the Inception Score (IS), also computed with PANNs. Cross-modal alignment is measured using ImageBind~\cite{ImageBind} for semantic alignment and DeSync~\cite{synchformer} for temporal alignment.

\noindent\textbf{Subjective Metrics.} For subjective evaluation, we conduct a user study involving 30 participants. We sample 20 videos from VGGSound and Countix-AV test sets and generate a 5-second audio clip for each video using each V2A method. Participants rate each generated audio clip on a 5-point scale in terms of overall audio quality (OAQ), semantic alignment (Sem.Align.) with the visual content, and temporal alignment (Temp.Align.) with the visual events. We report the mean $\pm$ standard deviation for each criterion to assess the perceived audio quality, semantic and temporal alignment. Please refer to the supplementary material for further details.

\subsection{Main Results}
\noindent\textbf{Objective Evaluation.}
We compare SCR-V2A with state-of-the-art V2A methods on VGGSound (in-domain) and CountixAV (out-of-domain), as shown in \Cref{tab:Tab1_objectiveEval}. On VGGSound, SCR-V2A achieves strong overall performance, obtaining the best FAD (1.07) and KLD (1.67), together with competitive FD (8.13), IS (13.81), and cross-modal alignment (IB: 29.95; DeSync: 0.680). Notably, Seeing\&Hearing achieves the highest IB by directly optimizing ImageBind similarity during inference, whereas SCR-V2A requires no metric-specific test-time optimization.
Under domain shift on CountixAV, SCR-V2A achieves the best FD (20.58), FAD (2.16), and KLD (2.13), while remaining competitive in IS (5.40), IB (33.17), and DeSync (0.632), indicating strong robustness and generalization. 
Moreover, SCR-V2A uses only 195M trainable parameters, substantially fewer than several large-scale competitors such as MMAudio (1.03B) and FoleyCrafter (1.22B), demonstrating a favorable balance between model capacity and generation performance.

\input{tableTex/Tab2_subjectiveEval}

\noindent\textbf{Subjective Evaluation.}
\Cref{tab:Tab2_subjectiveEval} summarizes the subjective evaluation results. SCR-V2A achieves the highest scores for overall audio quality and semantic alignment, outperforming all competing methods in both metrics. It also ranks second in temporal alignment, following MMAudio. Overall, SCR-V2A demonstrates strong subjective performance across all three evaluation criteria.

\noindent\textbf{Qualitative Results.}
\Cref{fig:Fig3_qualitativeResults} presents qualitative comparisons between the baselines and SCR-V2A. For the “playing double bass” video, SCR-V2A generates dense temporal patterns that correspond to the frequent yet subtle actions in the video. Its spectrogram contains repeated, structured energy peaks over time, suggesting greater sensitivity to rapid event dynamics rather than producing a static or overly smoothed response.

\input{figureTex/Fig4_temporalComposition}

\input{tableTex/Tab3_abl_visualPrior}
\subsection{Ablation Study}
\noindent\textbf{Comparison of Different Visual Priors.}
\Cref{tab:Tab3_abl_visualPrior} compares SCRs with bounding boxes,
segmentation masks, optical flow, and attention rollout. SCRs achieve
the best results across all metrics. Compared with the respective second-best results, it reduces FD, FAD, and KLD by 1.61, 1.28, and 0.55, respectively, while improving IS by 0.83 and IB by 3.44 percentage points and reducing DeSync by 0.123. These consistently superior results validate the effectiveness of the proposed SCR-based visual conditioning.

\input{tableTex/Tab4_abl_condition}

\noindent\textbf{Contribution of Conditioning Components.}
As shown in \Cref{tab:Tab4_abl_condition}, each conditioning modality improves the unconditional baseline, confirming that each provides useful information. Combining two modalities yields further gains, with RGB + SCR achieving the best FD, KLD, and DeSync among the paired variants, indicating strong complementarity between global visual appearance and sound-relevant spatiotemporal cues. Finally, integrating all three conditioning modalities yields the best performance across all six metrics. These results demonstrate that SCR-V2A effectively leverages their complementary information to improve overall generation quality and cross-modal alignment.

\subsection{Extended Features}

\noindent\textbf{Temporal Sound Event Composition.}
\Cref{fig:Fig4_temporalComposition} presents the ability of SCR-V2A to compose multiple sound events according to changes in visual content. The first example contains sound events occurring sequentially within the same video. As the sound-producing activity changes over time, SCR-V2A generates the corresponding events within their respective temporal intervals. The second example presents a more challenging setting, where video segments depicting different sound events are artificially concatenated into a composite sequence. The generated audio adapts to these transitions, producing sound events that correspond to the visual content of each segment. These examples suggest that SCR-based conditioning can capture sound-related semantic cues together with their temporal dynamics, enabling the model to organize multiple sound events into a temporally coherent audio sequence.

\noindent\textbf{Sound Source Awareness.}
\Cref{fig:Fig5_sourceAwareness} illustrates that SCR-V2A identifies visual evidence that is closely associated with the actual production of sound. As shown in the example, rough visual guidance used by existing methods broadly emphasize the visible sound-producing object and its surrounding appearance, without explicitly distinguishing which local regions are most relevant to the acoustic event. In contrast, our SCR guidance concentrates on fine-grained regions which correspond more closely to the actual sound-producing regions. Such localized cues provide spatially precise guidance beyond coarse object-level semantics, allowing SCR-V2A to focus on visual details that are informative of how the corresponding sound event is produced. This behavior suggests that SCRs provide the model with stronger awareness of sound-producing regions and their underlying visual interactions.

\input{figureTex/Fig5_sourceAwareness}

%% file: tableTex/Tab1_objectiveEval.tex
\begin{table*}[!htbp]
\centering
\caption{
Objective Evaluation. Comparison of SCR-V2A with baseline methods on the VGGSound (in-domain) and CountixAV (out-of-domain) test sets.
Note that Seeing\&Hearing~\cite{xing2024seeing} directly optimizes the IB score at test time and therefore achieves the highest IB score. 
Best results are shown in \textbf{bold}, and second-best results are \underline{underlined}.
}
\label{tab:Tab1_objectiveEval}
\resizebox{0.95\textwidth}{!}{\begin{tabular}{l c cccccc cccccc}
\toprule
\multirow{2}{*}{Method} &
\multirow{2}{*}{\#Params} &
\multicolumn{6}{c}{VGGSound} &
\multicolumn{6}{c}{CountixAV} \\
\cmidrule(lr){3-8} \cmidrule(lr){9-14}
 & & FD $\downarrow$ & FAD $\downarrow$ & KLD $\downarrow$ & IS $\uparrow$ & IB (\%) $\uparrow$ & DeSync $\downarrow$ 
 & FD $\downarrow$ & FAD $\downarrow$ & KLD $\downarrow$ & IS $\uparrow$ & IB (\%) $\uparrow$ & DeSync $\downarrow$ \\
\midrule
Diff-Foley \cite{luo2023diff}  & 859M 
&  25.60 & 7.99  & 3.17 & 9.85  & 19.87 & 0.884
&  42.11 & 9.82 & 3.36 & 5.36 & 21.13 & 0.853 \\ 

Seeing\&Hearing \cite{xing2024seeing}      & 415M 
&  27.70  & 5.40& 2.90 & 7.51 & \textbf{35.40} & 1.207 
& 35.72 & 8.98 & 2.65 & 3.26 & \textbf{34.95} & 1.198 \\

V2A-Mapper \cite{wang2024v2a}    & 229M 
&  9.69 & \underline{1.21} & 2.90 & 12.36 & 22.00 & 1.230
& \underline{21.43} & 3.99 & 3.07 & \textbf{5.46} & 18.41 & 1.210 \\ 

SS2A \cite{SS2A}     & 85M 
& 16.21 & 2.61 & 2.29 & 9.47 & 23.98 & 1.213
& 26.49 & 5.04 & 2.54 & 3.11 & 22.68 & 1.212\\ 

FoleyCrafter \cite{FoleyCrafter}     & 1.22B 
& 13.51 & 2.32 & 2.55 & 10.33 & 27.73 & 1.220
& 24.21 & 3.51 & 2.79 & 5.06 & 27.21 & 1.272\\ 

MMAudio \cite{mmaudio}          & 1.03B    
& \textbf{6.58} & 1.84 & \underline{1.93} & \textbf{16.88} & \underline{32.01} & \textbf{0.440}
& 23.07 & \underline{2.22} & \underline{2.39} & 5.09 & 32.07 & \textbf{0.371} \\ 

\midrule
\textbf{SCR-V2A (Ours)}  & 195M 
& \underline{8.13} & \textbf{1.07} & \textbf{1.67} & \underline{13.81} & 29.95 & \underline{0.680}
& \textbf{20.58} & \textbf{2.16} & \textbf{2.13} & \underline{5.40} & \underline{33.17} & \underline{0.632}\\

\bottomrule
\end{tabular}
}
\end{table*}

%% file: tableTex/Tab2_subjectiveEval.tex
\begin{table}[!htbp]
\centering
\caption{
Subjective Evaluation. The best and the second best results are shown in bold and by \underline{underlined}.
}
\label{tab:Tab2_subjectiveEval}
\resizebox{\columnwidth}{!}{\begin{tabular}{l cc cc cc}
\toprule
Method &
OAQ $\uparrow$ &
Sem.Align. $\uparrow$ &
Temporal.Align.$\uparrow$  \\
\midrule
See \& Hear              &  3.17 $\pm$ 1.21               &   3.63 $\pm$ 1.27               &   2.83 $\pm$ 1.21             \\
V2A-Mapper             &  3.23 $\pm$ 1.13               &   3.13 $\pm$ 1.28               &   2.73 $\pm$  1.23            \\
MMAudio                  &  \underline{3.80} $\pm$ 0.85   &   \underline{3.87}  $\pm$ 0.90  &   \textbf{3.87} $\pm$  0.94   \\
\midrule
\textbf{SCR-V2A (Ours)}  &  \textbf{3.93} $\pm$ 0.94       &   \textbf{4.10} $\pm$ 0.88      &   \underline{3.50} $\pm$ 0.97 \\    
\bottomrule
\end{tabular}
}
\end{table}

%% file: figureTex/Fig4_temporalComposition.tex
\begin{figure}[!hbtp]
  \centering
  \includegraphics[width=0.95\linewidth]{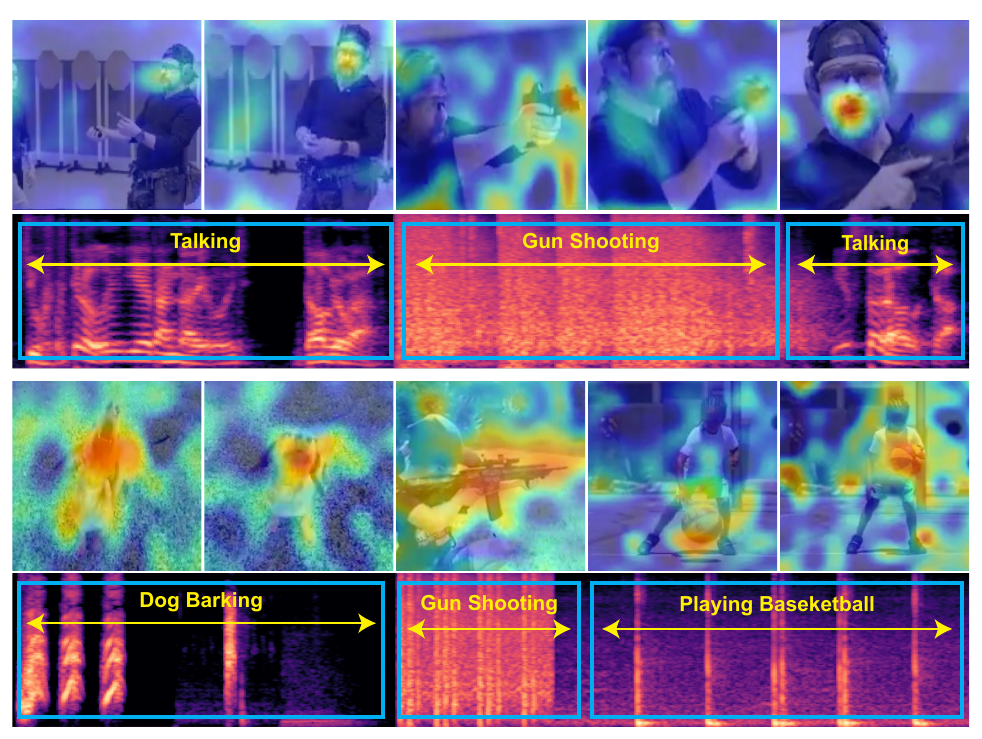}
  \Description{Two examples of temporally composed sound events. In the first sequence, SCR heatmaps shift between a person's face and a gun while the generated spectrogram is divided into talking, gun-shooting, and talking intervals. In the second sequence, heatmaps shift among a barking dog, a gun, and a basketball player, with the aligned spectrogram divided into dog-barking, gun-shooting, and basketball-playing intervals.}
  \caption{Temporal Sound Event Composition.}
  \label{fig:Fig4_temporalComposition}
\end{figure}

%% file: tableTex/Tab3_abl_visualPrior.tex
\begin{table}[!htbp]
\centering
\caption{
Comparison of Different Visual Priors. 
$^\ast$: extracted using YOLOv8~\cite{YOLOv8};
$^\dagger$: extracted using RAFT~\cite{RAFT}.
Best results are shown in \textbf{bold}, and second-best results are \underline{underlined}. 
}
\label{tab:Tab3_abl_visualPrior}
\resizebox{\columnwidth}{!}{\begin{tabular}{l cccccc}
\toprule
Visual Prior
& FD $\downarrow$ & FAD $\downarrow$ & KLD $\downarrow$ & IS $\uparrow$ & IB (\%) $\uparrow$ & DeSync $\downarrow$ \\

\midrule

Object BBox$^\ast$
& 14.36 & 3.48 & 2.97 & 11.80 & 23.45 & 1.168 \\

Object Mask$^\ast$
& 14.99 & 3.29 & 2.92 & 11.93 & 23.38 & 1.157\\

Optical Flow$^\dagger$
& 19.52 & 4.01 & 3.11 & 10.64 & 21.88 & \underline{0.803}\\

Attention Rollout
& \underline{9.74} & \underline{2.35} & \underline{2.22} & \underline{12.98} & \underline{26.51}  & 0.924\\

\midrule

\textbf{SCR} 
& \textbf{8.13} & \textbf{1.07} & \textbf{1.67} & \textbf{13.81} & \textbf{29.95} & \textbf{0.680}\\

\bottomrule
\end{tabular}
}
\end{table}

%% file: tableTex/Tab4_abl_condition.tex
\begin{table}[!htbp]
\centering
\caption{
Contribution of Conditioning Components.
Best and second-best results are shown in \textbf{bold} and \underline{underlined}. 
}
\label{tab:Tab4_abl_condition}
\resizebox{\columnwidth}{!}{\begin{tabular}{l cccccc}
\toprule
Condition
& FD $\downarrow$ & FAD $\downarrow$ & KLD $\downarrow$ & IS $\uparrow$ & IB (\%) $\uparrow$ & DeSync $\downarrow$ \\

\midrule

None & 33.20 & 9.31 & 4.82 & 7.14 & 10.62 & 1.842 \\

\midrule

Tag only & 25.85 & 5.31 & 4.08 & 10.31 & 21.06 & 1.536 \\

SCR only & 23.47 & 5.89 & 3.71 & 8.62 & 19.04 & 1.183 \\

RGB only & 22.74 & 5.76 & 3.49 & 9.27 & 20.82 & 1.284 \\
\midrule

Tag + SCR & 17.59 & 3.62 & 3.28 & 10.05 & 24.02 & 0.997 \\

Tag + RGB & 13.79 & \underline{1.62} & 2.97 & \underline{13.76} & \underline{27.18} & 1.174 \\

RGB + SCR & \underline{10.36} & 2.18 & \underline{2.35} & 11.82 & 26.57 & \underline{0.801} \\

\midrule

\textbf{SCR-V2A (Ours)} 
& \textbf{8.13} & \textbf{1.07} & \textbf{1.67} & \textbf{13.81} & \textbf{29.95} & \textbf{0.680}\\

\bottomrule
\end{tabular}
}
\end{table}

%% file: figureTex/Fig5_sourceAwareness.tex
\begin{figure}[!htbp]
  \centering
  \includegraphics[width=0.99\linewidth]{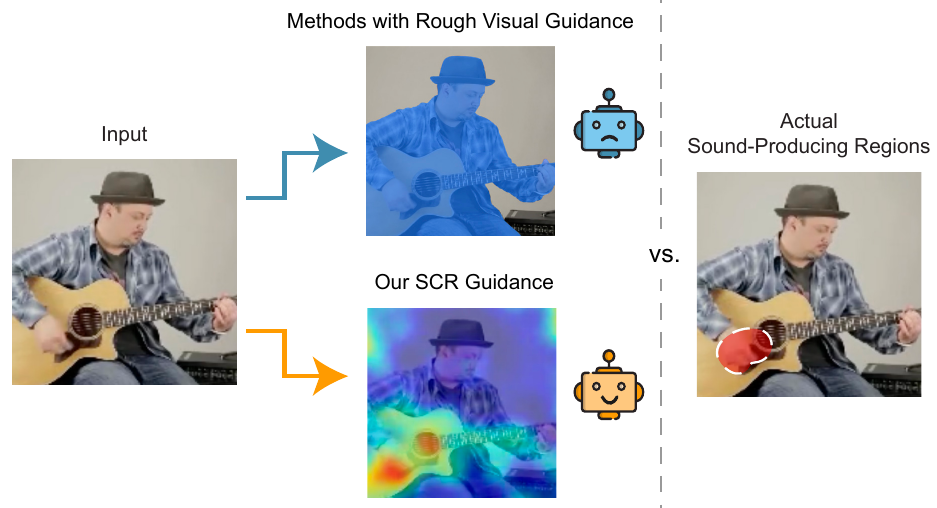}
  \Description{Comparison of rough visual guidance and SCR guidance for a video of a man playing an acoustic guitar. Rough guidance segments the entire person, whereas the SCR heatmap concentrates on the guitar and the player's hands. A reference image marks the guitar body and strumming hand as the actual sound-producing region.}
  \caption{Sound Source Awareness.}
  \label{fig:Fig5_sourceAwareness}
\end{figure}

%% file: sections/5_conc.tex
\section{Conclusion}
\label{sec:conclusion}

In this paper, we presented SCR-V2A, which explicitly localizes Semantic-Critical Regions and encodes them as fine-grained visual control for V2A generation. Experiments on in-domain and out-of-domain benchmarks demonstrate strong generation fidelity, cross-modal alignment, and generalization under multiple evaluation criteria, while requiring relatively few trainable parameters compared with several recent large-scale V2A methods. Analyses of temporal event composition and sound-source awareness further demonstrate the applicability and extensibility of SCR-based control to more flexible V2A generation settings.

%% file: sections/X_impl_details.tex
\section{Implementation Details}
\label{sec:X_impl_details}

\noindent\textbf{Data Preprocessing.}
Following \cite{jeong2025read}, we randomly sample 5-second video segments from VGGSound for training. 
During testing, we extract clips with durations ranging from 2 to 7 seconds for the VGGSound and CountixAV test sets. 
Video frames are uniformly sampled at 25 fps.
All audio signals are resampled to 16~kHz. 
Following the default configuration in \cite{liu2023audioldm}, we compute Mel-spectrograms using 64 Mel bins with a window length of 1024. 
The frequency range is set to $f_{min}=0$ and $f_{max}=8000$, with a hop size of 160 and an FFT size of 1024. 
For a fair comparison, since all baselines generate either 8-second \cite{luo2023diff,mmaudio} or 10-second \cite{xing2024seeing,wang2024v2a, SS2A, FoleyCrafter} audio for each test video, we use their default output durations and then extract the 5-second audio segment corresponding to the same video frames used by our method.

\noindent\textbf{Visual Encoder and Visual Prior Extraction.}
We adopt SynchFormer~\cite{synchformer} as the video encoder, trained on VGGSound~\cite{VGGSound}. 
Its backbone is based on MotionFormer~\cite{motionformer}, pre-trained on Something-Something v2~\cite{sth-sth} and further fine-tuned on VGGSound and AudioSet~\cite{AudioSet}, providing motion-sensitive and semantically rich representations. 
To derive the visual prior, we attach a lightweight classifier head on top of the visual encoder and apply Grad-CAM~\cite{GradCAM} to the last transformer block. 
The extractor is trained using AdamW, with a learning rate of $1\times10^{-5}$ for the backbone (with all layers frozen except the final block) and $1\times10^{-4}$ for the classifier head, under a binary cross-entropy objective. 
The labels used to train the extractor are derived from the official descriptions provided with VGGSound. 
During inference, the visual priors are computed based on the predicted class scores from the classifier head without access to ground-truth labels.

\noindent\textbf{Audio Generation Backbone.}
We build upon an AudioLDM-based latent diffusion model~\cite{liu2023audioldm,audioldm2} for audio generation. 
The ControlNet-style adapter~\cite{controlnet} are jointly optimized using AdamW with a base learning rate of $1\times10^{-5}$ and 2k warm-up steps. 
We follow the standard noise prediction objective for diffusion training. 
Classifier-free guidance is applied to the conditioning inputs with a dropout rate of 0.3, improving robustness under partially missing or noisy conditions.

\noindent\textbf{Training and Inference.}
We train SCR-V2A for 40 epochs on a single NVIDIA H100 GPU with a batch size of 64. 
During training, the visual prior is pre-computed using a separately trained extractor and kept frozen, together with the conditioning encoders, VAE, and vocoder. 
During inference, we employ the DDIM sampler with 200 steps to generate one audio sample per input. 
The output duration is set to 5 seconds, consistent with the training setup.

%% file: sections/X_expr_ratio.tex
\section{Effect of Different SCR Mixing Ratio}

\Cref{tab:Tab5_abl_diffRatio} evaluates the effect of varying the mixing ratio $\lambda$ in the visual prior conditioning. The intermediate setting of $\lambda=50\%$ achieves the best performance across all six metrics, with FD, FAD, KLD, IS, IB, and DeSync of 8.13, 1.07, 1.67, 13.81, 29.95, and 0.680, respectively. Both lower and higher ratios result in degraded performance, with particularly large drops observed at 70\% and 100\%. These results indicate that an intermediate mixing ratio provides the most effective visual prior conditioning, while overly emphasizing either SCR or RGB information leads to suboptimal performance.

\input{tableTex/Tab5_abl_diffRatio.tex}

\noindent \Cref{fig:Fig6_mixingRatios} visualizes the performance trends across different mixing ratios. All six metrics show a consistent pattern, with performance improving from 0\% to 50\% and degrading at higher ratios. The best performance is achieved at 50\% across all metrics, providing a clear view of the overall trend across the evaluated settings.

\input{figureTex/Fig6_mixingRatios.tex}

%% file: tableTex/Tab5_abl_diffRatio.tex
\begin{table}[!htbp]
\centering
\caption{
Effect of Mixing Ratio in Visual Prior Conditioning.
}
\label{tab:Tab5_abl_diffRatio}
\resizebox{\columnwidth}{!}{\begin{tabular}{l cccccc}
\toprule
$\lambda$
& FD $\downarrow$ & FAD $\downarrow$ & KLD $\downarrow$ & IS $\uparrow$ & IB (\%) $\uparrow$ & DeSync $\downarrow$ \\

\midrule

0\% (SCR only)
& 9.75 & 1.88 & 2.33 & 13.01 & 27.81 & 0.808 \\

30\%
& 8.82 & 1.57 & 2.02 & 13.68 & 29.12 & 0.699 \\

\textbf{50\%}
& \textbf{8.13} & \textbf{1.07} & \textbf{1.67} & \textbf{13.81} & \textbf{29.95} & \textbf{0.680}\\

70\%
& 10.70 & 2.11 & 2.51 & 12.11 & 25.10 & 0.991\\

100\% (RGB only)
& 14.43 & 3.37 & 3.10 & 11.90 & 23.30 & 1.170\\

\bottomrule
\end{tabular}
}
\end{table}

%% file: figureTex/Fig6_mixingRatios.tex
\begin{figure}[!htbp]
  \centering
  \includegraphics[width=0.98\linewidth]{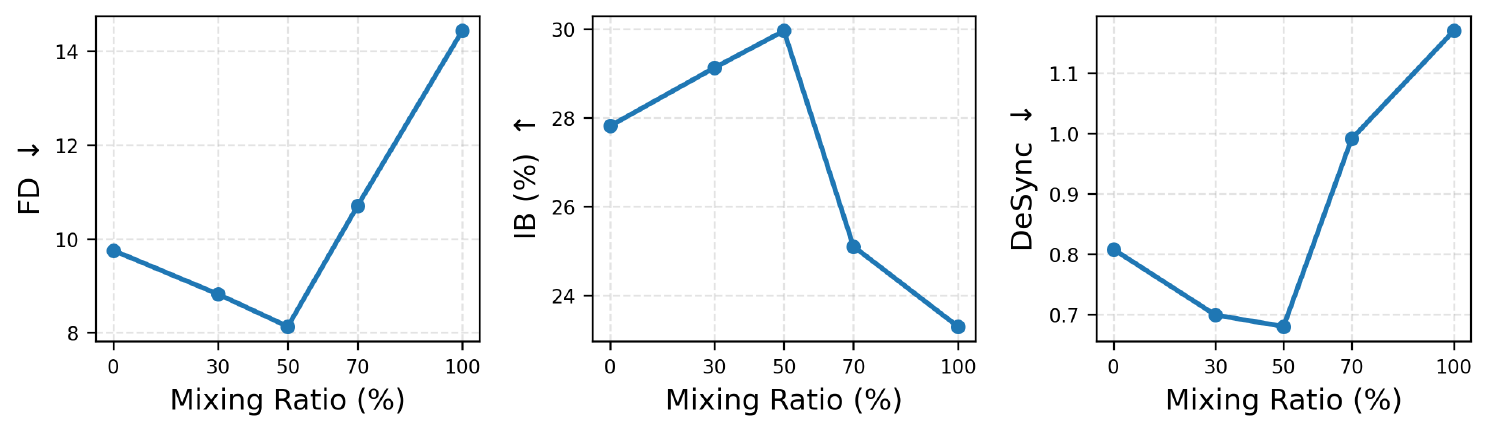}
  \Description{Three line charts plot FD, ImageBind score, and DeSync against SCR mixing ratios of 0, 30, 50, 70, and 100 percent. Performance is best at a 50 percent mixing ratio: FD and DeSync reach their minima while the ImageBind score reaches its maximum. All three metrics worsen when the ratio increases from 50 to 100 percent.}
  \caption{Performance Trends across Different Mixing Ratios.
}
\label{fig:Fig6_mixingRatios}
\end{figure}

%% file: sections/X_expr_negative_heatmap.tex
\input{figureTex/Fig7_humanEval}

\section{Effect of Negative SCR Heatmap Control}
\label{sec:X_expr_negative_heatmap}

To further examine whether the model relies on meaningful SCR guidance, we conduct an experiment using several negative heatmap variants as the control signal, as shown in Table~\ref{tab:Tab6_abl_negativeHeatmap}. Specifically, we consider:
\begin{itemize}
    \item \textbf{Uniformly Random Heatmaps}, where both spatial and temporal structures are destroyed by sampling values from a uniform distribution.
    \item \textbf{Shuffled Heatmaps}, where spatial locations are randomly permuted within each frame, preserving global activation statistics but disrupting localized patterns.
    \item \textbf{Inverted Heatmaps}, which emphasise background regions instead of semantically relevant sound sources by inverting the original heatmap values.
\end{itemize}
All three negative controls lead to substantial degradation across the evaluated metrics. The uniformly random variant increases FD from 8.13 to 18.34 and DeSync from 0.680 to 1.180, while reducing IB by nearly 7 percentage points. Shuffling the heatmaps results in further deterioration, particularly in FD and IB, whereas inversion produces the weakest overall performance, with FD reaching 30.15 and IB dropping to 17.54. The same overall trend is also observed in FAD, KLD, and DeSync, showing consistent degradation across distribution matching, audio quality, and cross-modal alignment.
These results indicate the importance of the localization structure in SCR conditioning. The larger degradation observed for shuffled and inverted heatmaps further highlights the contribution of the original SCR activation patterns to effective visual conditioning.

\input{tableTex/Tab6_abl_negativeHeatmap.tex}

%% file: figureTex/Fig7_humanEval.tex
\begin{figure*}[htbp]
  \centering

  \includegraphics[width=0.40\linewidth]{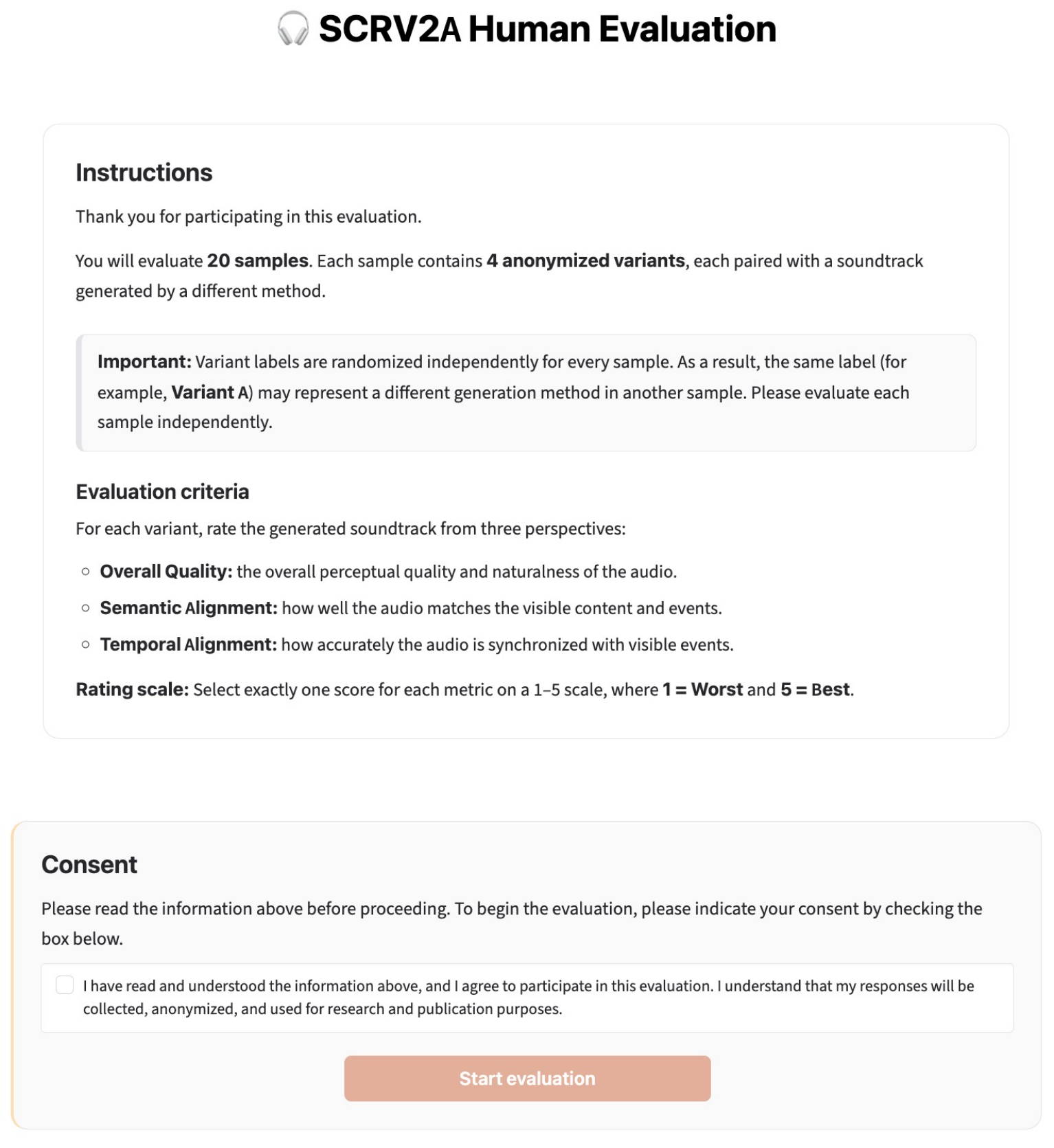}
  \hspace{0.03\linewidth}
  \includegraphics[width=0.42\linewidth]{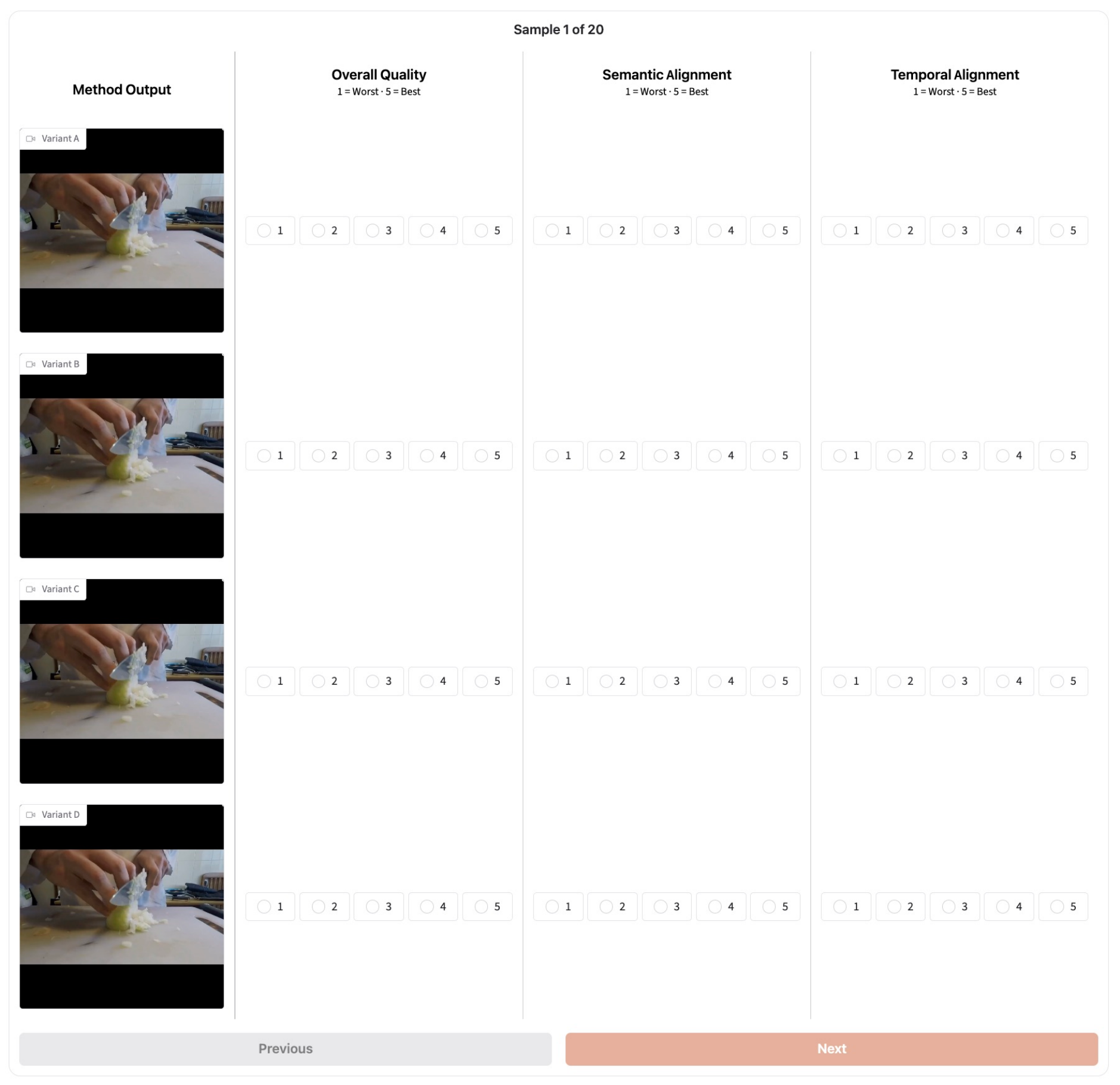}

  \vspace{0.5em}

  \includegraphics[width=0.42\linewidth]{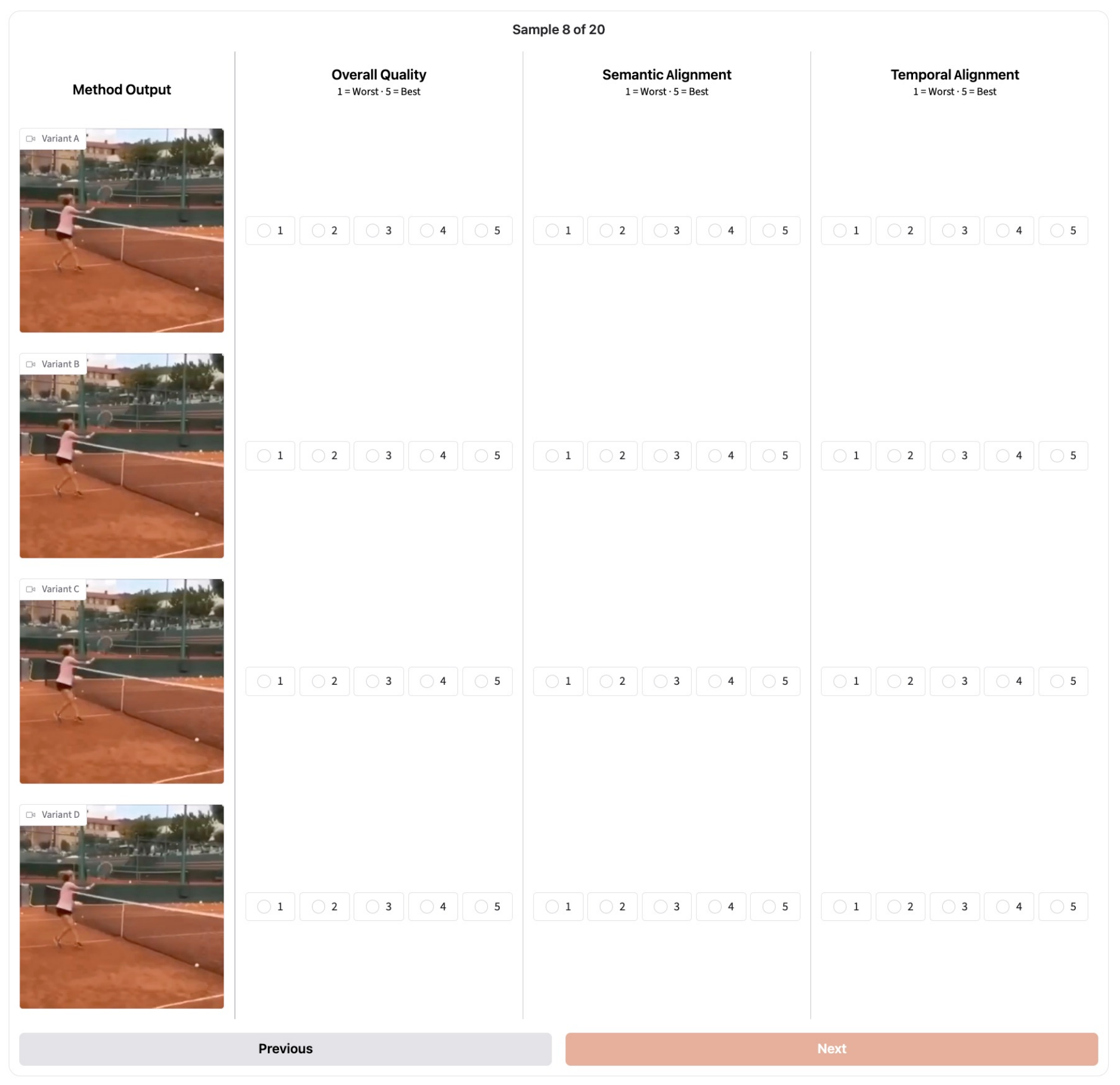}
  \hspace{0.03\linewidth}
  \includegraphics[width=0.42\linewidth]{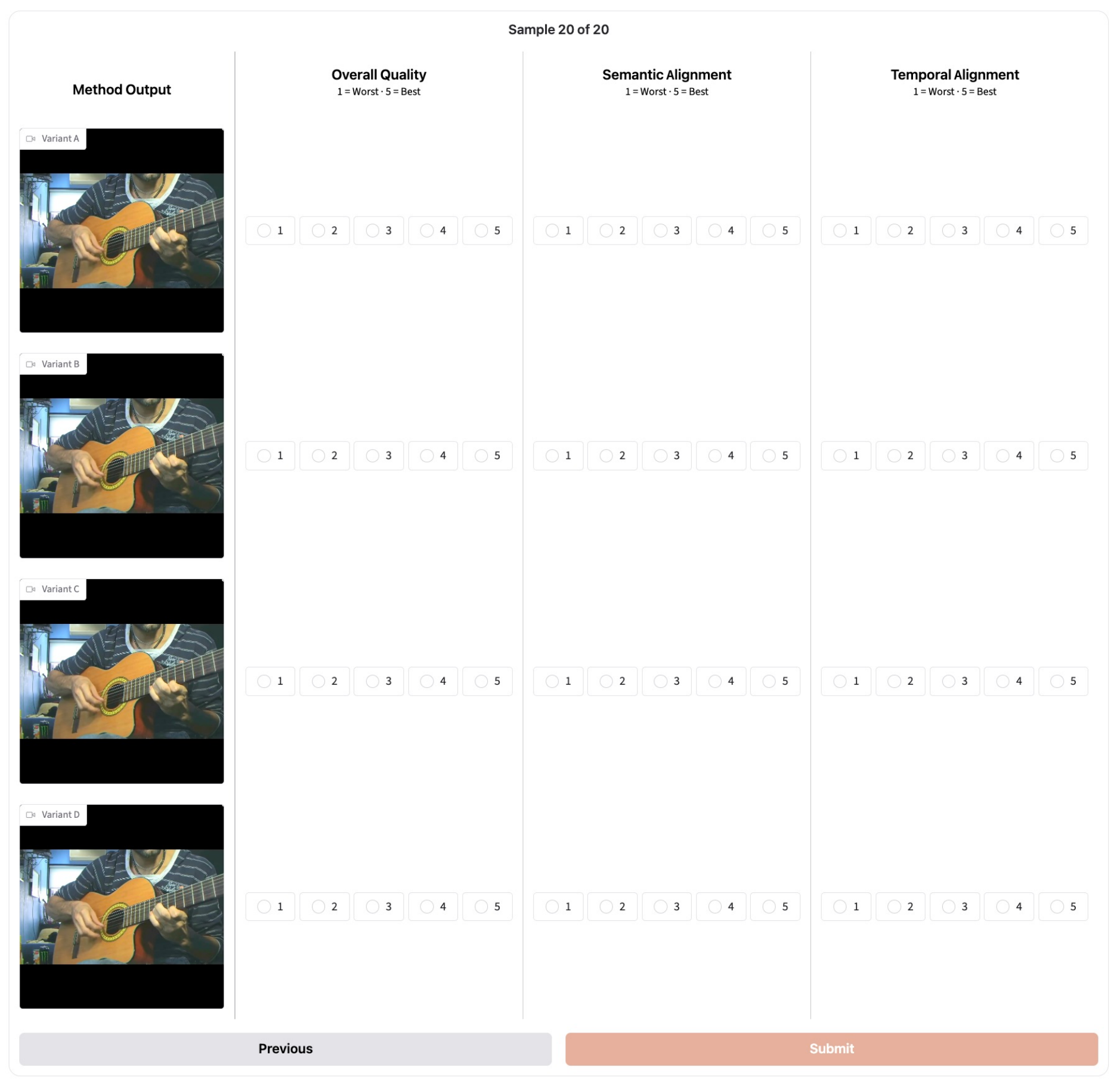}

  \Description{Four screenshots of the SCR-V2A human-evaluation website. The first shows instructions, consent, three evaluation criteria, and a one-to-five rating scale. The other screenshots show sample pages containing four anonymized video-and-audio variants, with separate one-to-five controls for overall quality, semantic alignment, and temporal alignment. Example videos depict food preparation, tennis, and guitar playing.}
  \caption{Screenshots of Subjective Evaluation Pages.}
  \label{fig:Fig7_humanEval}
\end{figure*}

%% file: tableTex/Tab6_abl_negativeHeatmap.tex
\begin{table}[!htbp]
\centering
\caption{
Effect of Negative Heatmap Control. 
}
\label{tab:Tab6_abl_negativeHeatmap}
\resizebox{\columnwidth}{!}{\begin{tabular}{l ccc c cc}
\toprule
\multirow{2}{*}{Model} &
\multicolumn{3}{c}{Distribution Matching} &
\multicolumn{1}{c}{Audio Quality} &
\multicolumn{2}{c}{Cross-modal Alignment}
\\
\cmidrule(lr){2-4} \cmidrule(lr){5-5} \cmidrule(lr){6-7}
 & FD $\downarrow$ & FAD $\downarrow$ & KLD $\downarrow$ & IS $\uparrow$ & IB (\%) $\uparrow$ & DeSync $\downarrow$ \\
\midrule

\textbf{SCR-V2A} 
& \textbf{8.13} & \textbf{1.07} & \textbf{1.67} & \textbf{13.81} & \textbf{29.95} & \textbf{0.680}\\

\quad w. uniformly random heatmaps & 18.34 & 4.02 & 3.11 & 10.63 & 23.07 & 1.180 \\

\quad w. shuffled heatmaps & 22.79 & 3.74 & 3.18 & 11.18 & 21.36 & 1.172 \\

\quad w. inverted heatmaps & 30.15 & 4.29 & 3.33 & 10.96 & 17.54 & 1.285 \\

\bottomrule
\end{tabular}
}
\end{table}

%% file: sections/X_human_eval.tex
\section{Subjective Evaluation}
\label{sec:human_eval}

We conduct a subjective evaluation using samples from the VGGSound and CountixAV test sets. We compare SCR-V2A with three strong baselines selected based on their performance in the objective evaluation:
\begin{itemize}
\item Seeing and Hearing~\cite{xing2024seeing}, which achieves the highest ImageBind score, indicating the strongest semantic alignment with the input videos.
\item V2A-Mapper~\cite{wang2024v2a}, which demonstrates strong audio quality, achieving the second-best FAD on VGGSound and FD on CountixAV, as well as the highest IS on CountixAV.
\item MMAudio~\cite{mmaudio}, which achieves the lowest DeSync score, indicating the strongest temporal alignment.
\end{itemize}

We sample 20 videos from the two test sets, comprising 12 samples from VGGSound and 8 from CountixAV. Each participant evaluates 80 video-audio variants in total (20 videos $\times$ 4 methods). For each video, the four variants are presented as a group, with their order independently randomized to mitigate ordering bias. Participants rate each variant along three dimensions using a five-point scale, ranging from 1 (worst) to 5 (best), according to the following criteria:
\begin{itemize}
\item \textbf{Overall Quality}: The overall perceptual quality and naturalness of the generated audio.
\item \textbf{Semantic Alignment}: The extent to which the generated audio matches the visible content and events.
\item \textbf{Temporal Alignment}: The extent to which the generated audio is synchronized with the visible events.
\end{itemize}

In total, we collected 2,400 ratings for each evaluation dimension from 30 participants. As reported earlier, SCR-V2A achieves the highest mean scores for overall audio quality and semantic alignment while ranking second in temporal alignment, demonstrating strong subjective performance across all three dimensions.

\noindent\textbf{Ethics Statement. }The evaluation was conducted through a dedicated web-based interface, as illustrated in \Cref{fig:Fig7_humanEval}. Before beginning the evaluation, participants were presented with the study instructions and asked to provide explicit informed consent by selecting a confirmation checkbox. Their responses were collected anonymously and used solely for research and publication purposes, without collecting sensitive personal information. 
The subjective evaluation was approved by the Human Research Ethics Committee of the University of Sydney (Approval No. 2023/HE000549).

%% file: sections/X_limitation.tex
\section{Limitations and Future Work}
\label{sec:limitation}
In this section, we discuss two key limitations of SCR-V2A and consider potential directions for future improvement.
First, the effectiveness of SCR conditioning largely depends on the recognition and localization quality of the SCR extractor. The qualitative examples presented throughout the paper suggest that the model can tolerate occasional low-salience activations on irrelevant regions without obvious degradation. However, incorrect semantic recognition may introduce persistent or substantially misleading spatial guidance and consequently degrade generation quality, as reflected in the negative heatmap control ablation in \Cref{tab:Tab6_abl_negativeHeatmap}. Improving the reliability of the SCR extractor therefore remains a promising direction for future work. More advanced visual encoders, localization techniques, or open-vocabulary recognition models could potentially enable more robust and generalizable SCR extraction.

Second, our current conditioning mechanism adopts a relatively simple and resource-efficient design. A ControlNet-style adapter is used to inject the conditioned visual features into the generation network, while SCRs and the original visual representation are combined through straightforward feature fusion. This design allows SCR guidance to be incorporated effectively without substantially increasing the complexity of the overall framework. Nevertheless, there remains room to explore richer conditioning strategies, both in how SCRs interact with the original visual features and in how the resulting representations are injected into the generation backbone. More advanced mechanisms, such as adaptive feature fusion, hierarchical integration, and more flexible feature injection, could be explored to make more effective use of SCR guidance while preserving the efficiency of the current design.